# CX_DB8:
# A queryable extractive summarizer and semantic search engine

**Allen Roush**
University of Oregon, USA

## Abstract

Competitive Debate's increasingly technical nature has left competitors looking for tools to accelerate evidence production. We find that the unique type of extractive summarization performed by competitive debaters – summarization with a bias towards a particular target meaning – can be performed using the latest innovations in unsupervised pre-trained text vectorization models. We introduce CX_DB8, a queryable word-level extractive summarizer and evidence creation framework, which allows for rapid, biasable summarization of arbitrarily sized texts. CX_DB8s usage of the embedding framework Flair means that as the underlying models improve, CX_DB8 will also improve. We observe that CX_DB8 also functions as a semantic search engine, and has application as a supplement to traditional "find" functionality in programs and webpages. CX_DB8 is currently used by competitive debaters and is made available to the public at https://github.com/Hellisotherpeople/CX_DB8.

## 1 Introduction

Extractive summarization is the task of automatically producing a summary of a text by deleting uninformative tokens. This is in contrast to abstractive summarization, which allows for deletion and replacement or insertion of tokens. One way of characterizing their difference is by using the highlighter vs pen analogy. Extractive Summarization is the process of highlighting a document to only include the most important or salient parts of the document. Abstractive Summarization is when one writes a completely new abstract based on the document. This abstract may include tokens which are not found in the document. CX_DB8 is an extractive summarization system.

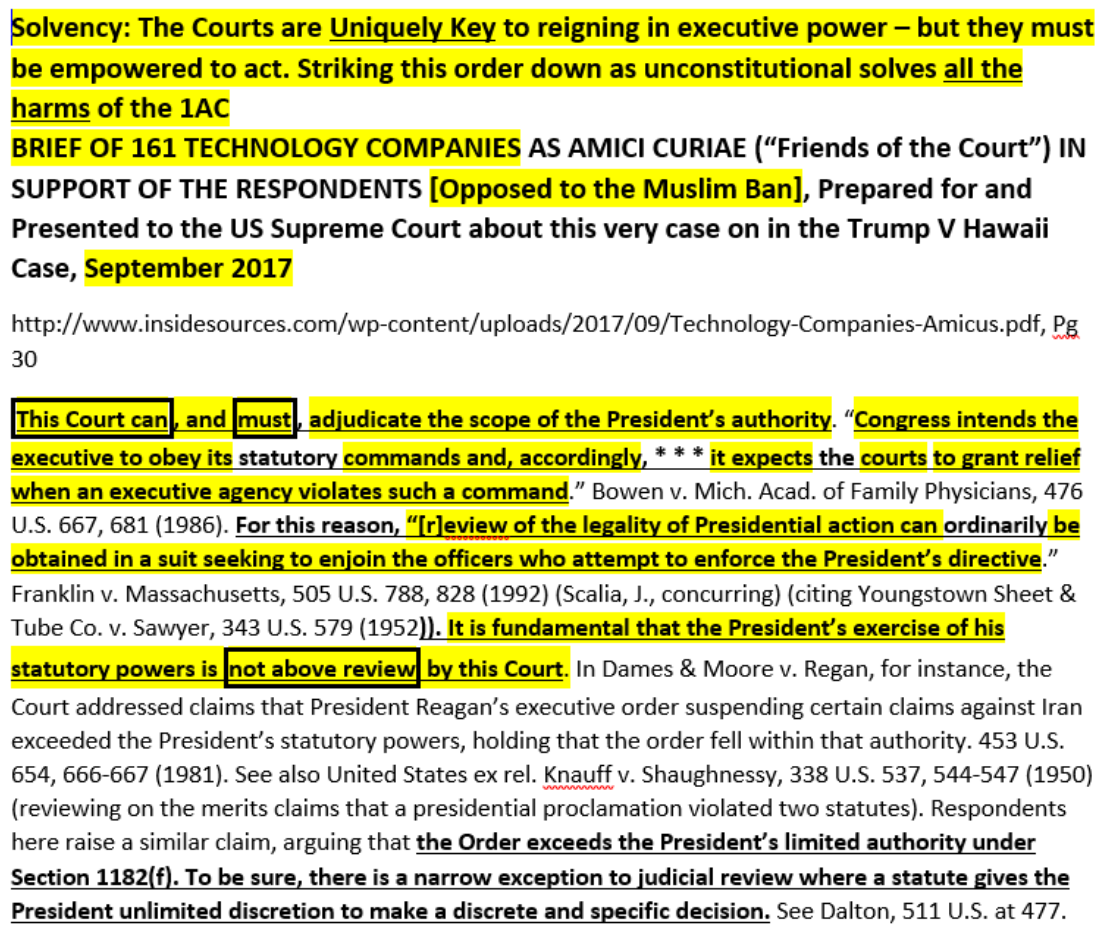

**Solvency: The Courts are Uniquely Key to reigning in executive power – but they must be empowered to act. Striking this order down as unconstitutional solves all the harms of the 1AC**
**BRIEF OF 161 TECHNOLOGY COMPANIES AS AMICI CURIAE ("Friends of the Court") IN SUPPORT OF THE RESPONDENTS [Opposed to the Muslim Ban], Prepared for and Presented to the US Supreme Court about this very case on in the Trump V Hawaii Case, September 2017**

http://www.insidesources.com/wp-content/uploads/2017/09/Technology-Companies-Amicus.pdf, Pg 30

**This Court can, and must, adjudicate the scope of the President's authority.** "**Congress intends the executive to obey its statutory commands and, accordingly, * * * it expects the courts to grant relief when an executive agency violates such a command**." Bowen v. Mich. Acad. of Family Physicians, 476 U.S. 667, 681 (1986). **For this reason, "[r]eview of the legality of Presidential action can ordinarily be obtained in a suit seeking to enjoin the officers who attempt to enforce the President's directive**." Franklin v. Massachusetts, 505 U.S. 788, 828 (1992) (Scalia, J., concurring) (citing Youngstown Sheet & Tube Co. v. Sawyer, 343 U.S. 579 (1952)). **It is fundamental that the President's exercise of his statutory powers is not above review by this Court**. In Dames & Moore v. Regan, for instance, the Court addressed claims that President Reagan's executive order suspending certain claims against Iran exceeded the President's statutory powers, holding that the order fell within that authority. 453 U.S. 654, 666-667 (1981). See also United States ex rel. Knauff v. Shaughnessy, 338 U.S. 537, 544-547 (1950) (reviewing on the merits claims that a presidential proclamation violated two statutes). Respondents here raise a similar claim, arguing that **the Order exceeds the President's limited authority under Section 1182(f). To be sure, there is a narrow exception to judicial review where a statute gives the President unlimited discretion to make a discrete and specific decision.** See Dalton, 511 U.S. at 477.

Figure 1: An example of a document used in the Competitive Debate community. The first three bolded and highlighted sentences are an abstract of the document. An extract of the document is made by highlighting all important text after the author and date.

Prior work in extractive summarization systems includes many recent advances made in NLP (Jadhav and Rajan, 2018; Zhao et al., 2018). Unfortunately, these systems suffer from some key limitations. Most of them do not utilize the latest in pre-trained word and character embeddings. Due to a focus on maximizing grammatical correctness, they only populate summaries with candidate text at the sentence level. They also have a focus on creating "faithful" summaries – e.g. they usually select sentences with the goal of taking the most similar sentences to the document as a whole. We find that documents can sometimes be "summarized to say" whatever a reader wants them to say.

CX_DB8's algorithm is inspired by how competitive debaters abstractively summarize documents, which presents their accounting of what the evidence says and how it supports their argument. After presenting this summarization, they recite an extractive summarization of that evidence. CX_DB8 imitates this process.

## 2 Related Work

Queryable summarization is not a new development. Yulianti et al., (2016) describe a system that concatenates tweets about a document alongside the document before summarizing. Users found the summaries of this system preferable to those of unbiased summaries. Azar et al., (2012) describes a query-based summarizer using Ensemble Noisy Auto-Encoders to select sentences. They test this system on emails using email subject lines as the query. Lierde and Chow (2019) describes a hypergraph-based system from summarization based on a query which tries to maximize coverage of the query by selecting the most semantically meaningful sentences via graph traversal. Chaudhari and Mattukoyya (2018) describes a tf-idf based document summarization model that biases towards specific words by incorporating a secondary "polarity" measuring model which selects sentences based on their sentiment. Notably, they anticipate the possibility of applications in businesses using biased summaries to create positive sounding testimonials from product buyers. They also anticipate the possibility of users "censor[ing] content in a graceful manner". Barve and Desai (2015) describes three query based extractive models for summarizing Sanskrit documents. Sanskrit was chosen due to all of the known Sanskrit texts in existence being digitized.

Most contemporary extractive summarizers focus on *sentence-level* document summarization. (Nallapati et al., 2016; Narayan et al., 2018; Shi et al., 2018). Even current word embedding based techniques for Extractive Summarization only work on the sentence level (Liu, 2019). Yet, word level extractive summarization systems do exist, and are sometimes called "Sentence Compression" systems (Filippova et al., 2015; Klerke et al., 2016). To our knowledge, none of these systems are queryable, nor do they utilize pre-trained word embeddings at any point in their training process.

Our algorithm is similar to the existing text rank algorithm (Mihalcea and Tarau, 2004) . TextRank can find keywords or to select sentences to form a summary. CX_DB8 blends a sliding word-window approach to computing word importances with the ranking mechanism described by Mihalcea and Tarau to generate its summaries.

Finally, we observe parallels between our work and semantic search engines. A Chrome extension called Fuzbal (Ilchenk, 2016) utilizes Word2Vec and GloVe embeddings to perform semantic searches of webpages. CX_DB8 performs a similar process on a document using a customizable set of embeddings. Fuzbal utilizes small pre-trained models in the interests of not impacting page render speed.

## 3 Backend Architecture

CX_DB8 underlying Architecture are unsupervised, pre-trained text vectorization models. Text vectorization is the process of converting text into an N dimensional vector. The vector properties of magnitude and direction allow for comparison between words using cosine distance. Some text vectorization models are trained to predict the next word given previous context, or are designed to predict the context words given a candidate word (Mikolov et al., 2013). More recent methods utilize newer neural network architectures like the Transformer and introduce techniques like Masking and bidirectionality (Devlin et al., 2018) for improved contextual disambiguation.

Work in the field of text embeddings is rapidly progressing. For this reason, CX_DB8 is powered by the text embedding framework Flair (Akbik and Blythe, 2018). Flair is chosen because it has a simple interface that allows a user to combine word or document embeddings together arbitrarily. The authors of Flair make it a point to quickly incorporate new text embedding models – meaning that CX_DB8's summarization capabilities will improve as the state-of-the-art models which power it improve.

Many summarization techniques utilize supervised techniques, but we consciously avoided them, as we desired the stronger generalization performance available from unsupervised models. Since unsupervised models are usually trained on massive corpuses, like Wikipedia or Common Crawl (Penninglon et al., 2014), they do not overfit as much to any particular topic or domain. Furthermore, they offer the user the ability to fine-tune the embedding models with a domain specific

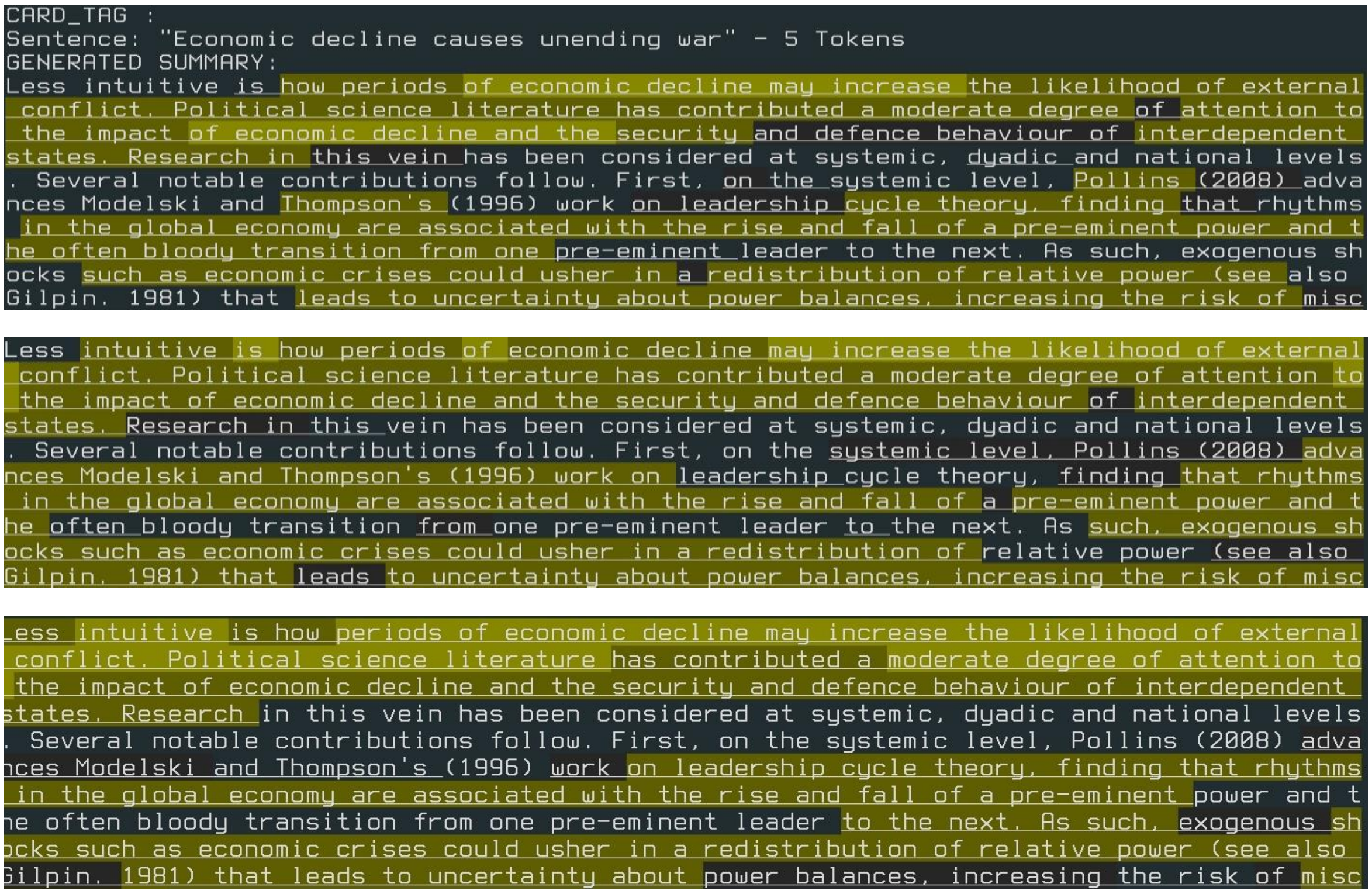

Figure 2: Comparison between different lengths of word window. Top is with length of 6, middle with length of 12, and the bottom with length of 20. Embeddings used: fastText trained on Wikipedia data, with the top 70% underlined and the top 65% highlighted

Bias query: “Economic decline causes unending war”

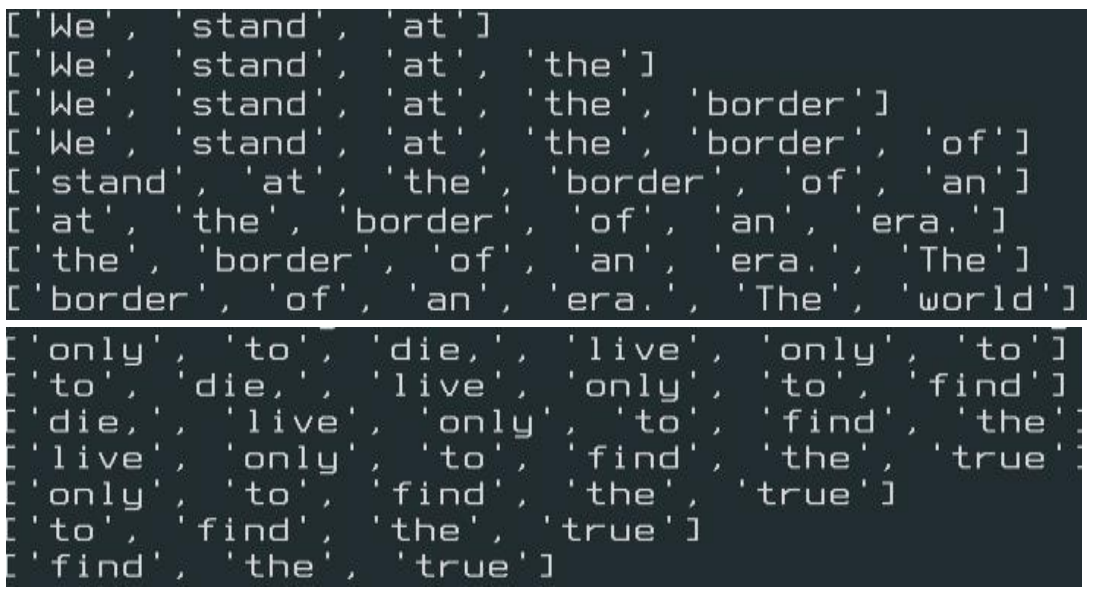

Figure 3: Scaling word-windows from the beginning and end of a document with a word window size of 6. The window for the first word “we” is the first 3 words of the summary. The window for the second word “stand” adds the next word to the word-window, CX_DB8 appends additional words to the list as it slides the window through the text, eventually creating 6 word bidirectional word windows through the middle of the text, and reducing back to 3 for the final word, “true”

unlabeled corpus, rather than requiring labeled text data.

## 4 Algorithm Overview

### 4.1 Word Windows

CX_DB8 can generate a user specified length summary from a document by computing the cosine similarity between a vectorized representation of a user-inputted query and each words corresponding *word-window* in the document. This produces a scaler for each word, which corresponds to the similarity of this word (and context) to the query. Figure 2 displays the difference in summarization as the word window is lengthened. We observe that as the word window increases, the summaries tend to include longer runs of words, roughly proportional to the size of the word window.

Compared to sentence level summarization (which assigns scalers to sentences rather than words), this technique trades off a significant amount of grammatical correctness. For the purposes of the competitive debate community, such tradeoffs are worth the benefits in scalable summarization. Furthermore, it improves the capacity for a biased summary to piece together a

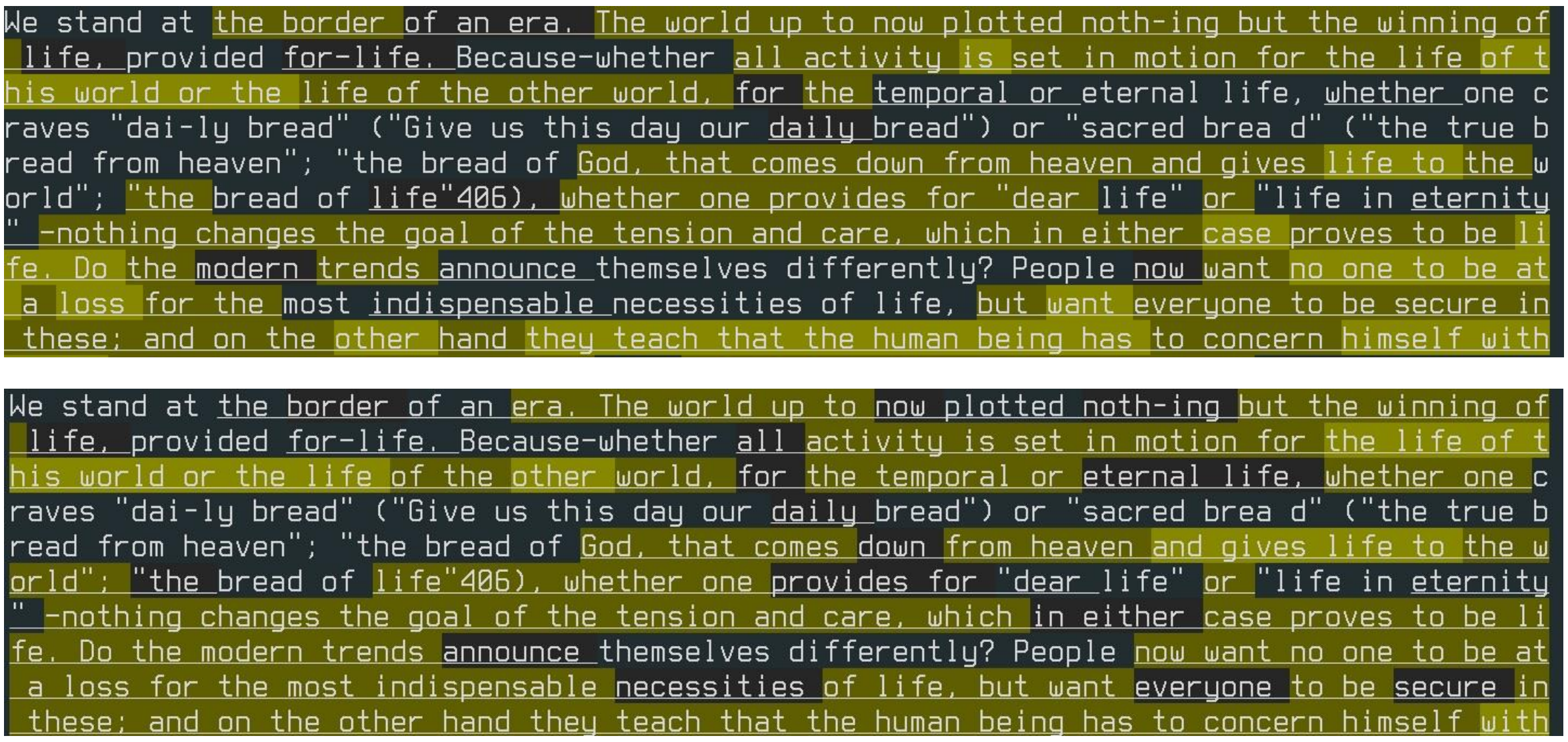

Figure 4: Unbiased summary (top) vs biased summary (bottom) Hyperparamaters chosen: fastText trained on Wikipedia data, with the top 70% underlined and the top 65% highlighted

Bias query: "The moderns seek enjoyment of life rather than staying alive"

meaning that was not intended by the original documents authors.

The Scaling Word Window as described in Figure 3 was chosen for its perceived natural correspondence to how a human may read a text, as well as its usage in the pretraining of the models that CX_DB8 uses.

### 4.2 Bias Query

When executing CX_DB8, it first asks a user to input a bias query. This can be a single word (for semantic search), a sentence (or a "tagline" as debaters call it), or even a whole different document. Users who wish to generate an unbiased summary can enter "-1" and CX_DB8 will set the query to be the document to be summarized. Each word is vectorized and the bias vector is computed by mean pooling each word vector in the query.

Figure 4 highlights the differences in summarization between unbiased and biased summaries. In theory, a summary can be generated that "says what a user wants it to say", but in practice, the nature of word embeddings makes this challenging. Because word embeddings predict context, The sentences "I love ice cream" and "I hate ice cream" will rank as being extremely similar to each other, and a query looking for the ills of ice cream may accidentally end up including all of the information about how good ice cream is. Still, we observe the queryability feature to be an extremely useful tool for guiding the algorithm's summary creation.

### 4.3 Summarization

During the main execution loop, the user enters the documents that they wish to summarize. CX_DB8 takes input in through the system's standard input stream and a user indicates that they are done entering their document by pressing ctrl-d.

Once the user has entered their document, CX_DB8 works as follows: For each word in the source document, the vectorized representation of the word window is computed. CX_DB8 then computes the cosine similarity between the query vector and each word window vector. CX_DB8 prompts the user to specify the percentage of the document that they want underlined and highlighted ahead of time. The words with the highest similarity are selected to be included in the summary. CX_DB8 also allows for a user to re-underline or re-highlight a document if they want a different sized summary or a summary with different settings.

### 4.4 Pooling

CX_DB8 inherits all hyperparamaters available in Flair's DocumentPoolEmbeddings class. Any set of word vectors available through Flair, as well as custom user-trained models, can be selected to power CX_DB8. For instance, a user could leverage fine-tuned word2vec, GloVe, fastText, and BERT models to summarize a document. Flair concatenates the embeddings together seamlessly.

Computing the vector representation of a multi-word string is usually done by averaging each word or character vector, though Flair makes it possible to take the maximum or minimum of each vector. We utilize average word pooling by default.

As CX_DB8 was written originally for the competitive debate community, it asks the user if more summaries are to be produced. If the user indicates that they are finished, it produces a word document in a similar format as the evidence shown in Figure 1. It also outputs this summary using standard output and Sty [1] for dynamic highlighting in a terminal.

## 5 Use Cases

This section describes current domains where CX_DB8 is successfully used, and proposes further use cases and extensions.

### 5.1 Competitive Speech and Debate

The original idea and raison d'etre for this tool was to assist in the creation of evidence for users who compete in American Style Cross Examination Debate. This style of debate is characterized by its length, extreme technical style, its heavy reliance on evidence, and its annual topic. Many competitors painstakingly spend hundreds of hours researching, and preparing evidence in the format shown in Figure 1. CX_DB8's name pays homage to this community.

When this tool was introduced into the competitive debate community, it caused mostly positive reactions. Many were impressed with the customizability and speed of summarization, but some believe that the automation of evidence production risks incentivizing competitors to not properly read research that they might include as a core part of their debate case. The ideal use case for this tool is to quickly summarize documents, which were found immediately prior to an important speech. As there is extremely limited preparation time, any competitor who utilizes CX_DB8 can have a significant advantage over their competition.

[1] https://github.com/feluxe/sty

### 5.2 Semantic Search

Some users of CX_DB8 noted the similarity that this model has to a semantic powered "find" or "search" tool. Typing a query like "Policies" on a politician's web page may easily guide a user to their list of policies, even if the webpage chooses different but semantically similar words to describe their policy page. Fuzbal proved that there is a significant group of users who would find such a feature useful if it were built into a web browser.

One especially interesting use case of CX_DB8 is in information retrieval. A user mentioned that they were trying to use CX_DB8 in tandem with embeddings trained on pubmed abstracts to process medical documents about cancer and search for experimental or novel treatments for their family. We strongly support these kind of efforts and hope that tools like CX_DB8 can be useful to those who need to quickly parse large amounts of information with semantic understanding

### 5.3 Future Work

While many in the debate community are satisfied with grammatically incorrect summaries, several users have requested a sentence level summarization mode. We plan to add this feature in a future update. Furthermore, there are exciting possibilities related to trying to incorporate further context into the ranking mechanism, such as by concatenating sparse and wide embedding models like tf-idf alongside current deep embedding models, or by fine tuning text vector models on debate corpuses.

## 6 Conclusion

We presented CX_DB8, a queryable, word-level extractive summarization system designed for a specific domain. This tool leverages state-of-the-art pre-trained language models to generate its summaries. We explain the underlying architecture behind CX_DB8, muse about potential use cases, and serve as a call to action for more work in the field of semantic search. Our system is utilized extensively within the competitive speech and debate community, and is made available to the public on github.